# Dynamic Programming for Structured Continuous Markov Decision Problems

**Zhengzhu Feng**
Department of Computer Science
University of Massachusetts
Amherst, MA 01003-4610
fengzz@cs.umass.edu

**Richard Dearden*, Nicolas Meuleau†, Richard Washington***
NASA Ames Research Center, Mail Stop 269-3
Moffet Field, CA 94035-1000
{dearden, nmeuleau, richw}@email.arc.nasa.gov

## Abstract

We describe an approach for exploiting structure in Markov Decision Processes with continuous state variables. At each step of the dynamic programming, the state space is dynamically partitioned into regions where the value function is the same throughout the region. We first describe the algorithm for piecewise constant representations. We then extend it to piecewise linear representations, using techniques from POMDPs to represent and reason about linear surfaces efficiently. We show that for complex, structured problems, our approach exploits the natural structure so that optimal solutions can be computed efficiently.

## 1 INTRODUCTION

Markov Decision Processes (MDPs) have been adopted as a framework for much recent research in decision-theoretic planning. Classic dynamic programming algorithms solve MDPs in time polynomial in the size of the state space. However, the size of the state space is usually very large in practice. For systems modeled with a set of propositional state variables, the state space grows exponentially with the number of variables. This problem becomes even more important for MDPs with continuous state-spaces, which we will refer to as *general state-space MDPs* (GSSMDPs) by analogy with general state-space Markov chains (Gilks, Richardson, & Spiegelhalter 1996). If the continuous space is discretized to find a solution, the discretization causes yet another level of exponential blow up. This "curse of dimensionality" has limited the use of the MDP framework, and overcoming it has become an important topic of research.

In discrete MDPs, model-minimization (Dean & Givan 1997) techniques have been used with considerable success to limit this state explosion problem. Algorithms such as SPI (Boutilier, Dearden, & Goldszmidt 2000) and SPUDD (Hoey *et al.* 1999) operate by identifying regions of the state-space that have the same value under the optimal policy, and treating those regions as a single state for the purposes of dynamic programming. Although they do not guarantee to avoid the combinatorial explosion in all cases, they increase the range of application of MDP algorithms to a wide class of real problems.

In this paper we extend this state aggregation to continuous problems. Figure 1 shows the optimal value from the initial state of a typical Mars rover problem as a function of two continuous variables: the time and energy remaining (Bresina *et al.* 2002). The shape of this value function is characteristic of the rover domain, as well as other domains featuring a finite set of goals with positive utility and resource constraints. Such a value function features a set of humps and plateaus, each of them representing a region of the state space where a particular goal (or set of goals) can be reached. The sharpness of a hump or plateau reflects the uncertainty attached to the actions leading to this goal. Morevover, constraints on the minimal level of resource required to start some actions (Bresina *et al.* 2002) introduce abrupt cuts in the regions. It results in a function with vast plateau regions where the expected reward is nearly constant. These correspond regions of the state space where the optimal policy tree is the same, and the probability distribution on future history induced by this optimal policy is nearly constant. The goal of this work is to exploit such structure by grouping together states belonging to the same plateau, while reserving a fine discretization for the regions of the state space where it is the most useful (such as the curved hump where there is more time and energy available).

We will show that for certain subclasses of GSSMDPs, optimal solution can be obtained efficiently by exploiting the structure in the problem to perform dynamic programming at far fewer points than a naive approach would. The approach we will describe is restricted to MDPs with piecewise constant or piecewise linear reward functions, and

* Research Institute for Advanced Computer Science.
† QSS Group Inc.

Figure 1: Value function in (Bresina *et al.* 2002)

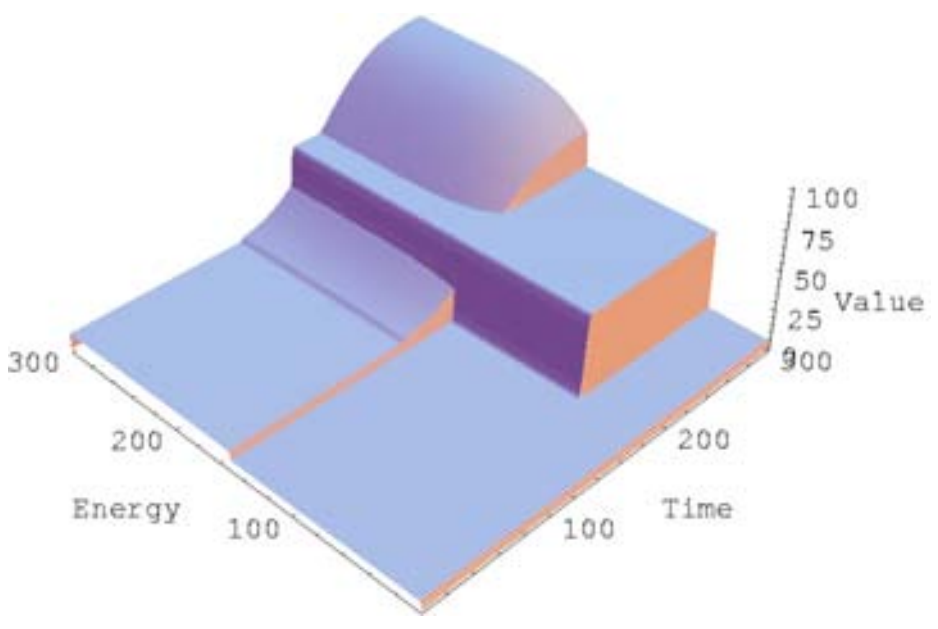

more significantly, to MDPs with discrete transition functions. This means that for any state and action, a finite set of states can be reached with non-zero probability. These restrictions ensure that the final optimal value function found by our algorithms belong to well-behaved families. In the case of a GSSMDP with discrete transition functions and piecewise-constant reward function, the optimal value function is guaranteed to be piecewise constant as well. Similarly, the use of piecewise-linear reward functions ensures a piecewise-linear value function.

The restriction to discrete transition functions is a particularly strong one. However, we can approximate MDPs with continuous transition functions by an appropriately fine discretization of the transition function. This provides an attractive alternative to function approximation approaches (Gordon 1995; Boyan & Moore 1995) in that it approximates the model but then solves the approximate model exactly, rather than finding an approximate value function for the original model. This has the advantage that the effect of the approximation can be much more easily quantified. It also contrasts with the naive approach that consists of discretizing the state space regardless of the relevance of the partition introduced. Instead, we discretize the action outcomes and deduce a partition of the state space from it. For ease of exposition, we will assume that any approximation has already been performed in the modeling process, and we will refer to finding optimal policies and value functions below, even when the model has been approximated.

Given these assumptions, the algorithm we describe will produce as its output a partition of the state-space in which each element of the partition consists of a region where the optimal value function is constant (for a piecewise-constant problem) or is the maximum of a set of linear functions (for a piecewise linear problem). This is achieved using representations of computational geometry to store state-space partitions, and POMDP techniques to compute Bellman back-ups in the piecewise linear case. On problems that exhibit structure, the algorithm computes the optimal value function substantially faster than a naive discretization of the state space, even when it must discretize more finely than the naive approach in some regions. The reduction in the number of Bellman backups performed to compute the optimal value function more than offsets the additional cost of maintaining the structured representation. Although this technique does not guarantee an exponential reduction of the complexity of solving continuous problems by discretizing them, it allows solving in a few minutes instances of the planetary rover problem that required in the order of one day of computation before.

## 2 NOTATION

We adopt a standard MDP model with a continuous state space: $\{\mathbf{X}, A, R, T\}$. $\mathbf{X}$ is a vector of continuous state variables $\langle X_1, \ldots, X_d \rangle$. In addition we assume the value of the variables are all in the range $[0, 1)$, so the state space is the unit square $[0, 1)^d$. We use $\mathbf{x} \in [0, 1)^d$ to refer to a particular state. $A$ is a finite set of actions. $R$ is the reward model: $R_a(\mathbf{x})$ is the reward for taking action $a$ in state $\mathbf{x}$. $T$ is the transition model. Following (Boyan & Littman 2000), we allow both *relative* and *absolute* transitions. A relative transition is expressed as $T_a(\mathbf{x} + \delta\mathbf{x}, \mathbf{x})$, the probability that the state is shifted by $\delta\mathbf{x}$ relative to $\mathbf{x}$. An absolute transtion is expressed as $T_a(\mathbf{x}', \mathbf{x})$, the probability that the resulting state is $\mathbf{x}'$. We generally refer to the finite set of possible resulting states from taking action $a$ in state $\mathbf{x}$ as the *outcomes*, denoted $\Delta^a_\mathbf{x}$. For relative outcomes, an element $\delta_i \in \Delta^a_\mathbf{x}$ is a pair $(\delta\mathbf{x}, p)$, where $p$ is the probability of that outcome. Similarly for absolute outcomes, $\delta_i$ is a pair $(\mathbf{x}', p)$.

We are interested in maximizing the expected total reward of a finite-horizon plan. The Bellman Equation is:

$$V^{n+1}(\mathbf{x}) = \max_{a \in A} \{R_a(\mathbf{x}) + \sum_{\mathbf{x}' \in \Delta^a_\mathbf{x}} T_a(\mathbf{x}', \mathbf{x}) V^n(\mathbf{x})\} \quad (1)$$

where $V^n(\mathbf{x})$ is the value function over the horizon of $n$ time-steps and $V^0(\mathbf{x}) = 0$.

## 3 STRUCTURAL ASSUMPTIONS AND REPRESENTATION

The structure that we exploit in this paper consists of partitioning the continuous state space into discrete regions, each of which can be treated as a single entity. In particular, we consider (hyper-)rectangular partitions of the state space $[0, 1)^d$. We will use the term “rectangle” or “region” instead of “hyper-rectangle” for brevity, and we will discuss examples from a 2-dimensional state space, but the formalism holds for arbitrary number of dimensions.

The important property of the models is that they are closed under the Bellman backup in Equation 1. There are many models that satisfy this property; as we said above, we consider piecewise constant and piecewise linear models here, and describe each of them in more detail in the following subsections.

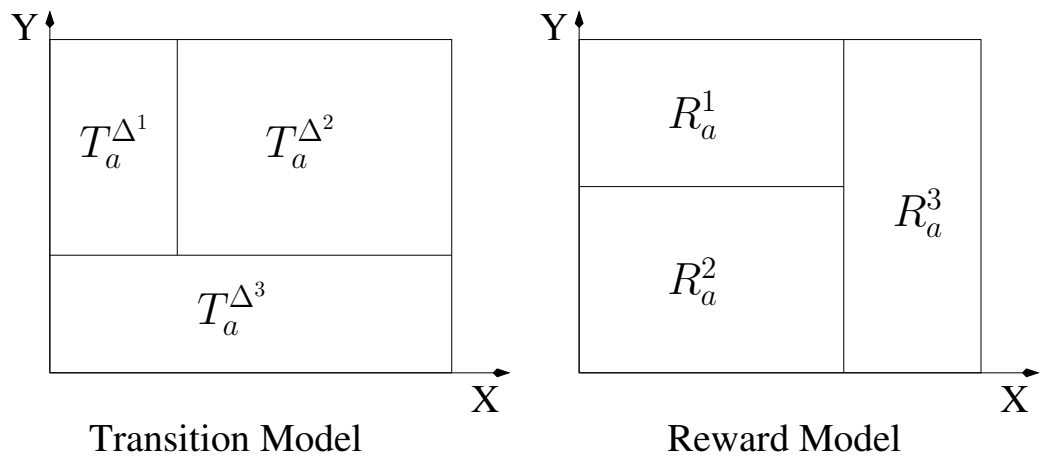

Figure 2: Rectangular piecewise constant models

### 3.1 PIECEWISE CONSTANT STRUCTURE

In this section, we assume the transition and reward model are *rectangular piecewise constant* (RPWC).

**Definition 1 Rectangular Partition** *A rectangular partition of the state space* $[0,1)^d$ *is a finite set of rectangles* $\boxplus = \{\Box_1, \Box_2, \ldots, \Box_k\}$*, where* $\Box_i = \prod [X_i^{low}, X_i^{high})$*, such that* $\bigcup_{1 \le i \le k} \Box_i = [0,1)^d$*, and* $\Box_i \cap \Box_j = \phi$ *iff* $i \neq j$*.*

**Definition 2 RPWC function** *A function* $f : [0,1)^d \rightarrow \mathbb{O}$ *is RPWC, if there exists a rectangular partition* $\boxplus = \{\Box_1, \Box_2, \ldots, \Box_k\}$ *such that* $\forall i, 1 \le i \le k$*, and* $\forall \mathbf{x}, \mathbf{y} \in \Box_i$*,* $f(\mathbf{x}) = f(\mathbf{y})$*.*

Note that the set $\mathbb{O}$ that a RPWC function maps to can either be the set of real numbers $\mathbb{R}$, in the case of the reward model, or the set of all possible outcome sets $\Delta$, in the case of the transition model.

As shown in Figure 2, the state space is partitioned into rectangular regions. For each action, the outcome set and the probability distribution over it are the same for all states inside a region of the transition model partition. We will use $\Delta_{\Box}^a$ to refer to the outcome set associated with a rectangle $\Box$ and an action $a$. In the case of a relative outcome set, for a region $\Box$,

$$\forall_{\mathbf{x},\mathbf{y} \in \Box} T_a(\mathbf{x} + \delta\mathbf{x}, \mathbf{x}) = T_a(\mathbf{y} + \delta\mathbf{x}, \mathbf{y}).$$

Thus a relative outcome can be seen as *shifting* a region. An absolute outcome maps states in a region to a single state $\mathbf{z}$:

$$\forall_{\mathbf{x},\mathbf{y} \in \Box} T_a(\mathbf{z}, \mathbf{x}) = T_a(\mathbf{z}, \mathbf{y}).$$

We will concentrate on the relative transition models, since they are more interesting from a formal and algorithmic standpoint. We will mention implications of absolute models where necessary.

For a specific action $a$, the transition model is represented by a partition $\boxplus_T^a$, and for each rectangle $\Box \in \boxplus_T^a$, a set of relative outcomes $\Delta_{\Box}^a$ together with a probability distribution over it. Similarily, the rewards $R_a$ are constant in each region. The reward model is represented by a rectangular partition $\boxplus_R^a$ and for each rectangle $\Box$ a constant $R_{\Box}$ representing the reward. Note that the partitions for the transition and reward model of an action need not be the same.

Applying RPWC assumptions to the standard model described in the previous section results in an MDP **M1** = $\{\mathbf{X}, A, T_{\boxplus}, R_{\boxplus}\}$, where $T_{\boxplus}$ and $R_{\boxplus}$ are RPWC transition and reward models as described above. We can show that

**Theorem 1** *For MDP* **M1***, if* $V^n$ *is RPWC, then* $V^{n+1}$ *computed by the Bellman backup (Equation 1) is also RPWC.*

Since we can represent a RPWC function exactly using a set of rectangles, this theorem enables us to carry out the Bellman backup exactly, assuming the initial value function is RPWC. States belonging to the same region of the value function: (i) have the same optimal policy; (ii) generate the same probability distribution on future history, in terms of actions performed, rewards received, and value function regions traversed under this optimal policy; and thus, (iii) have the same value.

**Dynamic Programming for M1** We now describe the Bellman backup procedure for the RPWC model. We first show how to compute the summation in Equation 1, which we denote as $\sigma_a$:

$$\sigma_a := \sum_{\mathbf{x}' \in \Delta_{\mathbf{x}}^a} T_a(\mathbf{x}', \mathbf{x}) V^n(\mathbf{x}')$$

We construct a partition for $\sigma_a$ by projecting the partition defined by the transition model of action $a$, namely $\boxplus_T^a$, onto the partition defined by $V^n$, using Procedure $\sigma_a$ listed in Figure 4. As an example, Figure 3 shows the subdividing process for a single rectangle $\Box \in \boxplus_T^a$. There are two relative outcomes for this action if taken in $\Box$, namely $\delta_1$ with probability 0.2 and $\delta_2$ with probability 0.8. For each outcome, we compute the new position of rectangle $\Box$, and intersect it with the partition of $V^n$. The result is then multiplied by the probability of the outcome. Finally, the results of all outcomes are intersected and the summation is computed within each sub-region of the intersection.

Note that this process assumes relative outcomes. For absolute outcomes, we need to modify step 1(a) so that the region $\Box_j$ is not subdivided, and is assigned the value of the outcome state in $V^n$ multiplied by the outcome probability.

The remainder of the Bellman backup involves adding the reward and performing the max over all possible actions. The full algorithm is listed as Procedure *Bellman backup* in Figure 4. In the whole process, a rectangle is further sub-divided only when necessary during the process of intersecting two partitions.

**KD-Tree Representation** For our implementation, we use kd-trees (Friedman, Bentley, & Finkel 1977) to store

Figure 3: Computing $\sigma_a$ for **M1**

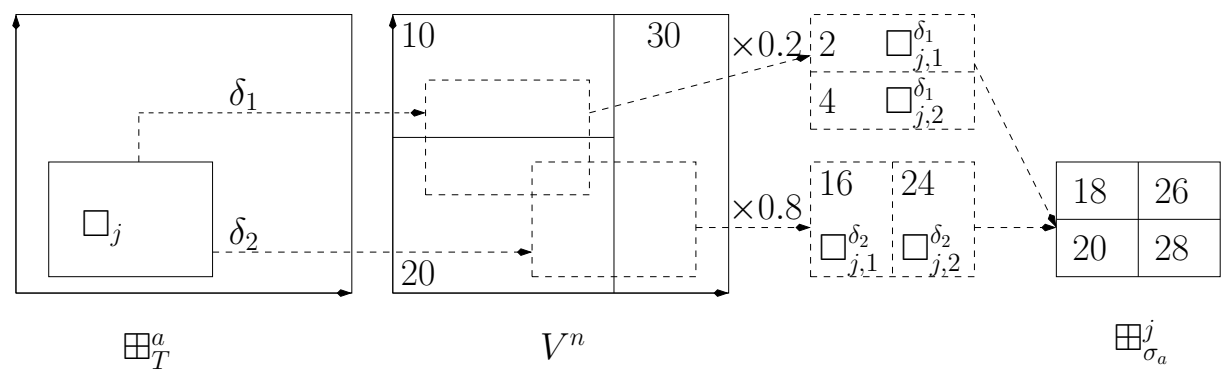

Figure 4: Dynamic Programming for **M1**

**Procedure** $\sigma_a$

1. For each region $\square_j$ in $\boxplus^a_T$
   (a) For each outcome $\delta_i \in \Delta^a_{\square_j}$
      i. Compute the region $\square^{\delta_i}_j$ resulting from shifting $\square_j$ by the relative outcome $\delta_i$.
      ii. Intersect the shifted region $\square^{\delta_i}_j$ with the partition of $V^n$, producing sub-regions $\square^{\delta_i}_{j,k}$
      iii. Assign to each sub-region $\square^{\delta_i}_{j,k}$ the value of the corresponding region of $V^n$ multiplied by the probability of the outcome $\delta_i$.
   (b) Intersect all the shifted regions from all of the outcomes, producing partition $\boxplus^j_{\sigma_a}$.
   (c) Assign to each of the regions in partition $\boxplus^j_{\sigma_a}$ the sum of the values of the corresponding sub-regions $\square^{\delta_i}_{j,k}$.
2. Assemble the final partition: $\boxplus_{\sigma_a} = \cup_j \boxplus^j_{\sigma_a}$.

**Procedure *Bellman backup***

1. Compute partition $\boxplus_{\sigma_a}$ for all $a$ using **Procedure** $\sigma_a$.
2. For each action $a$
   (a) Intersect partition $\boxplus_{\sigma_a}$ with $\boxplus^a_R$ to get partition $\boxplus_{Q_a}$.
   (b) The value of each region in $\boxplus_{Q_a}$ is computed by summing the values of the corresponding regions of $\boxplus_{\sigma_a}$ and $\boxplus^a_R$.
3. The partitions $\boxplus_{Q_a}$ of all actions are intersected, producing $\boxplus_{V^{n+1}}$.
4. The value of each region in $\boxplus_{V^{n+1}}$ is computed as the max of each of the corresponding regions in all the partitions $\boxplus_{Q_a}$.

and manipulate the rectangluar partitions. A kd-tree is a multidimensional generalization of the binary tree in which space is recursively split by hyper-planes orthogonal to one of the $k$ axes. Note that the partition induced by a kd-tree may contain unnecessary spliting of regions in a RPWC function. On the other hand, the intersection operation, which is the main computation of the whole algorithm, can be performed efficiently using algorithms such as (Naylor, Amanatides, & Thibault 1990) on kd-trees. Notably, these algorithms treat different number of dimensions in a uniform way, and have complexity that only depend on the size of the kd-trees. We omit details here for space reason.

Actions that have a relative effect on some variable tend to cause the partition of the value function to get finer as the horizon increases, which can affect the efficiency of the algorithm. For this reason, it can be necessary to implement a merging mechanism to unify neighboring regions with the same value. Value-based region merging breaks some of the properties of the model. Notably, states belonging to the same region no longer necessarily generate the same probability distribution over trajectories. However, this does not affect the outcome of the algorithm, while allowing computational savings. If these savings appear insufficient for a particular application, one can further reduce the complexity of the algorithm by merging pieces with similar value, trading quality of solution for computation time.

Merging based solely on value can break the RPWC property, resulting in partitions that can not be represented by kd-trees. For our implementation, we performed the merging taking into account both the value and the structure of the kd-tree representation, by performing a depth-first traversal of the kd-tree, and merging the leaf-nodes of the kd-tree if they have the same value. This way the kd-tree representation is maintained throughout the merging process.

### 3.2 PIECEWISE LINEAR AND CONVEX STRUCTURE

In this section, we extend the model **M1** by allowing more complex reward structures so that richer domains can be modeled. For example, to take into account the lighting of a rock from the sun in a rover problem, the value of taking a picture could vary linearly with time of the day. To model such structures, we extend the RPWC reward model to a *rectangular piecewise linear and convex* (RPWLC) reward model.

**Definition 3 PWLC** *A function $f$ over a region $\square$ is piecewise linear and convex, if there exists a finite set of linear functions $L = \{l_i | l_i(\mathbf{x}) = \mathbf{A}_i\mathbf{x} + \mathbf{B}_i\}$ such that $\forall \mathbf{x} \in \square, f(\mathbf{x}) = \max_{l_i \in L} l_i(\mathbf{x})$.*

We will generally use $L$ to refer to both a set of linear functions and the PWLC function that it represents.

**Definition 4 RPWLC** *A function $f : [0,1)^d \rightarrow \mathbb{R}$ is RPWLC if 1) there exists a rectangle partition $\boxplus = \{\square_1, \ldots, \square_k\}$; and 2) $\forall i, 1 \leq i \leq k$, there exists a PWLC function $L_i$ such that $\forall \mathbf{x} \in \square_i, f(\mathbf{x}) = L_i(\mathbf{x})$.*

This representation allows $R$ and $V^i$ for a region of the partition to be the maximum of a set of linear functions, rather than a single function as in **M1**. We allow this because a Bellman backup will create non-rectilinear regions when it performs the maximization step on linear functions.

We will refer to the model with the RPWLC assumption as **M2** $= \{\mathbf{X}, A, T_{\boxplus}, R^L_{\boxplus}\}$, where $R^L_{\boxplus}$ is the RPWLC reward model as defined above. Often, the reward function may be of a simpler form (a single linear function per region), but the resulting value function will remain RPWLC. Note that the transition model remains RPWC in **M2**.

The PWLC structure is a feature of *partially observable* MDP (POMDP) models (see, for example, (Cassandra, Littman, & Zhang 1997)). Thus, the RPWLC assumption allows us to adapt some existing results and practices in the POMDP literature. First we can show that,

**Theorem 2** *If $V^n$ is RPWLC, then $V^{n+1}$ as computed by the Bellman backup in Equation 1 is also RPWLC.*

As in the RPWC model **M1**, the operations during the Bellman backup for **M2** are intersection, summation, and max. The intersection is identical. The difference between **M1** and **M2** arises from the fact that for the value functions in **M2**, each rectangle contains a set of linear functions rather than a single scalar value. In particular, we need to perform *addition* and *maximization* between two sets of linear functions over the same rectangle. These operations are well defined in the POMDP literature. In particular, the addition of two sets of linear functions $L_1$ and $L_2$ is carried out by the *cross-sum* operator, and the maximization be carried out by the *union* operator:

**Definition 5 PWLC cross-sum and union** *Let $f_1$ and $f_2$ be two PWLC functions represented by $L_1$ and $L_2$, then the function $g(\mathbf{x}) = f_1(\mathbf{x}) + f_2(\mathbf{x})$ can be represented by the cross-sum of $L_1$ and $L_2$, defined as $L_1 \oplus L_2 := \{(l_i + l_j) | l_i \in L_1, l_j \in L_2\}$, and the function $h(\mathbf{x}) = \max\{f_1(\mathbf{x}), f_2(\mathbf{x})\}$ can be represented by the union set $L_1 \cup L_2$.*

Pruning is performed to remove dominated linear functions. Maximization over the remaining linear functions is used to determine the value and policy of any point in the state space. This considerably improves the efficiency of the algorithm, and the combination of pruning and maximization over the remaining linear functions achieves the maximization part of the Bellman equation. As in POMDPs, dominance is computed by solving a linear program. Many POMDP algorithms differ only in the way the pruning is carried out. Our implementation for RPWCL GSSMDPs prunes every intermediate function as in Incremental Pruning (Cassandra, Littman, & Zhang 1997).

To adapt the algorithms presented above so that they support the **M2** model, we make the following changes: In step (1)(c) of Procedure $\sigma_a$ and step (2)(b) of Procedure *Bellman backup*, we change sum of values to cross-sum of linear functions. In step (4) of Procedure *Bellman backup*, we change max of values to union of linear functions. Then we add pruning after step (1)(c) of Procedure $\sigma_a$ and after steps (2)(b) and (4) of Procedure *Bellman backup*.

# 4 MIXED DISCRETE-CONTINUOUS MODEL

In this section, we further extend the model to include discrete state components, which is the case of the Mars rover domain that motivated this work (Bresina *et al.* 2002). The new model is defined as **M3**= $\{S, \mathbf{X}, A, T_{\boxplus}, R^L_{\boxplus}\}$, where $S$ is a set of discrete states. The full state space is the product $S \times \mathbf{X}$. We will use $(s, \mathbf{x})$ to refer to a specific state in the state space. $T_a(s', \mathbf{x}', s, \mathbf{x})$ is the probability of reaching $(s', \mathbf{x}')$ if action $a$ is taken in state $(s, \mathbf{x})$. We will generally define the transition model by a a marginal probability distribution on the arrival discrete state: $T^d_a(s', s, \mathbf{x})$, and a conditional distribution over the continuous space $T^c_a(\mathbf{x}', s', s, \mathbf{x})$ given the arrival discrete state. As in **M1** and **M2**, the conditional distribution $T^c_a$ is assumed to be RPWC. For reward, we will define a function $R^s_a(\mathbf{x})$ for each action and discrete state pair, and assume that all $R^s_a$ are RPWLC. We represent the value function over the full state space using a set of functions $V := \{V_s(\mathbf{x}) | s \in S\}$. The Bellman backup for **M3** can be performed as:

$$V^{n+1}_s(\mathbf{x}) = \max_{a \in A}\{R^s_a(\mathbf{x}) + \sum_{s'} T^d_a(s'|s, \mathbf{x}) \sigma^{s'}_a\}$$

$$\sigma^{s'}_a = \sum_{\mathbf{x}'} T^c_a(\mathbf{x}', s', s, \mathbf{x}) V^n_{s'}(\mathbf{x}')$$

We can show that,

**Theorem 3** *If $V^n_s$ is RPWLC for $\forall s \in S$, then $V^{n+1}_s$ is also RPWLC for $\forall s \in S$.*

Algorithmically, the addition of discrete states changes the backup procedure by adding a loop over the discrete states to the computation described in the previous section.

# 5 EXPERIMENTAL RESULTS

This section depicts preliminary results obtained by solving prototype ("toy") instances of the Mars rover domain. Encouraged by these results, we are currently integrating the techniques presented here in an integrated architecture for planning and execution that allows solving real size instances of the problem (cf. section 7). This system uses other acceleration techniques and lies beyond the scope of the paper.

We tested our algorithms on a Mars rover domain adapted from (Bresina *et al.* 2002). The domain contains a "primary" plan, which consists of approaching a target point, digging the soil, backing up, and taking spectral images of the area. All these actions consume time and battery according to different Gaussian distributions. There are two potential branch to the primary plan: The first branch is to replace the spectral imaging with a high-resolution camera imaging, which in general consumes more time and en-

Figure 5: Results on the three sets of test problems from Mars rover domain. Times are measured in seconds, and the base of log is 2.

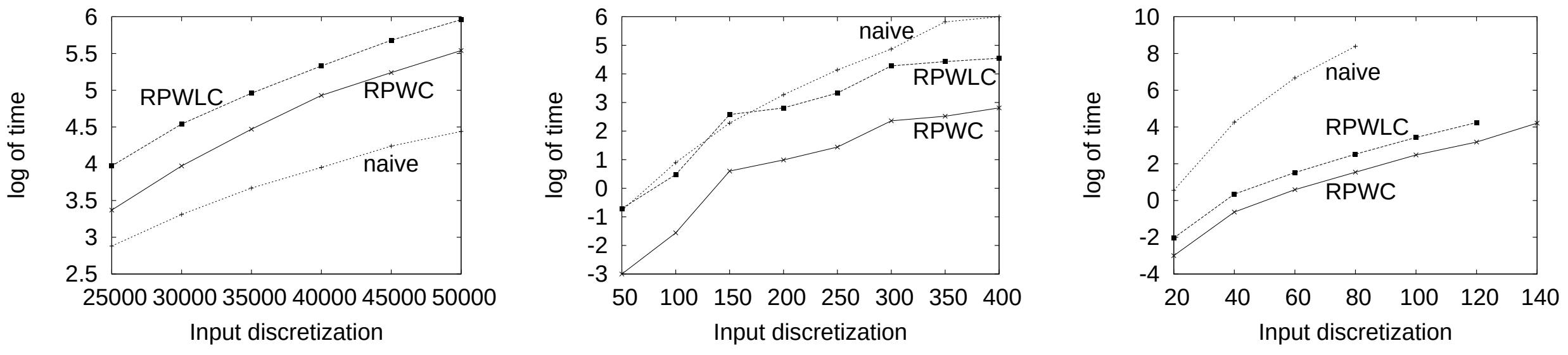

(a) Result on problem set 1D. (b) Result on problem set 2D. (c) Result on problem set 3D.

ergy. The second branch is to replace the digging-backing-imaging plan with a simple low-resolution imaging, and then perform onboard image analysis.

We model the domain with 11 discrete states representing different stages of the rover exploration process. We vary the number of continuous variables, which model different types of resources, from 1 to 3, creating three sets of test problems, referred to as 1D, 2D and 3D, respectively. In the original domain, action effects on the resources are modeled with continuous probability distributions. In our model, these continuous distributions are discretized. The resolution of the discretization is the independent variable in our experiments. For each resolution, we create two versions of the problem, one with constant rewards (RPWC representation), the other with rewards that are linear functions of the continuous variables (RPWLC representation). We compare the performance of our algorithm on each of these two representations with a naive algorithm that discretizes the value function using the same resolution used in the input discretization.

Figure 5 shows the results. The $X$-axis shows the input discretization resolution on each continuous variable. The $Y$-axis shows the elapsed run-time of the different algorithms, on a logarithmic scale. Note that for the naive approach, the run-time of the two versions (RPWC and RPWLC) of the problem are largely the same, because after the discretization of the value function, the identical amount of computation is carried out for both problems. Thus only the result on the RPWC problem is plotted.

As the figures show, our algorithm is slower than the naive approach for all the 1D problems. The overhead of dealing with the complex data structure exceeds the savings gained from it for the simple version of the problems. For the 2D problems, the RPWC model outperforms the naive approach. For lower input resolutions (from 50 to 150), the RPWLC model performs similarly to the naive approach. However, it is considerably faster for higher resolution problems. For the 3D problems, the difference between the naive approach and our approaches is more dramatic. In particular, the naive approach did not finish after 3 hours for problems with resolution greater than 80.

These results shows that our algorithm can scale better as the number of continuous variables increases. The improved scalability results from exploiting the specific problem structure to avoid unnecessary discretization of the value function. Figure 6 shows the resulting value function on a specific discrete state of a problem in the 2D set with RPWLC rewards, and an input discretization of 25 on each dimension. The left side shows the actual function, and the right side shows the corresponding partition over the continuous space. As we can see, fine discretization is only applied to the upper right region. Approximately 70% of the space is treated exactly with only a small number of regions. In contrast, the naive approach discretizes the entire space evenly, expending a large amount of computation on areas that are, in fact, from the same linear function. Our algorithm avoids this computation by treating large regions as a single state.

Two features in Figure 6 deserve further analysis. Firstly, although the input is only discretized with a resolution of 25, the resulting partition has considerably more discretization points, albeit all concentrated at the upper right region. This is because the initial partitions defining the transition and reward models are not necessarily aligned with the input discretization, so a finer partition is needed to represent the optimal value function. The naive approach doesn't make the additional distinctions so it misses details of the value function. The complexity grows over the course of the dynamic programming, so early iterations use coarser partitions, providing another saving over the naive approach.

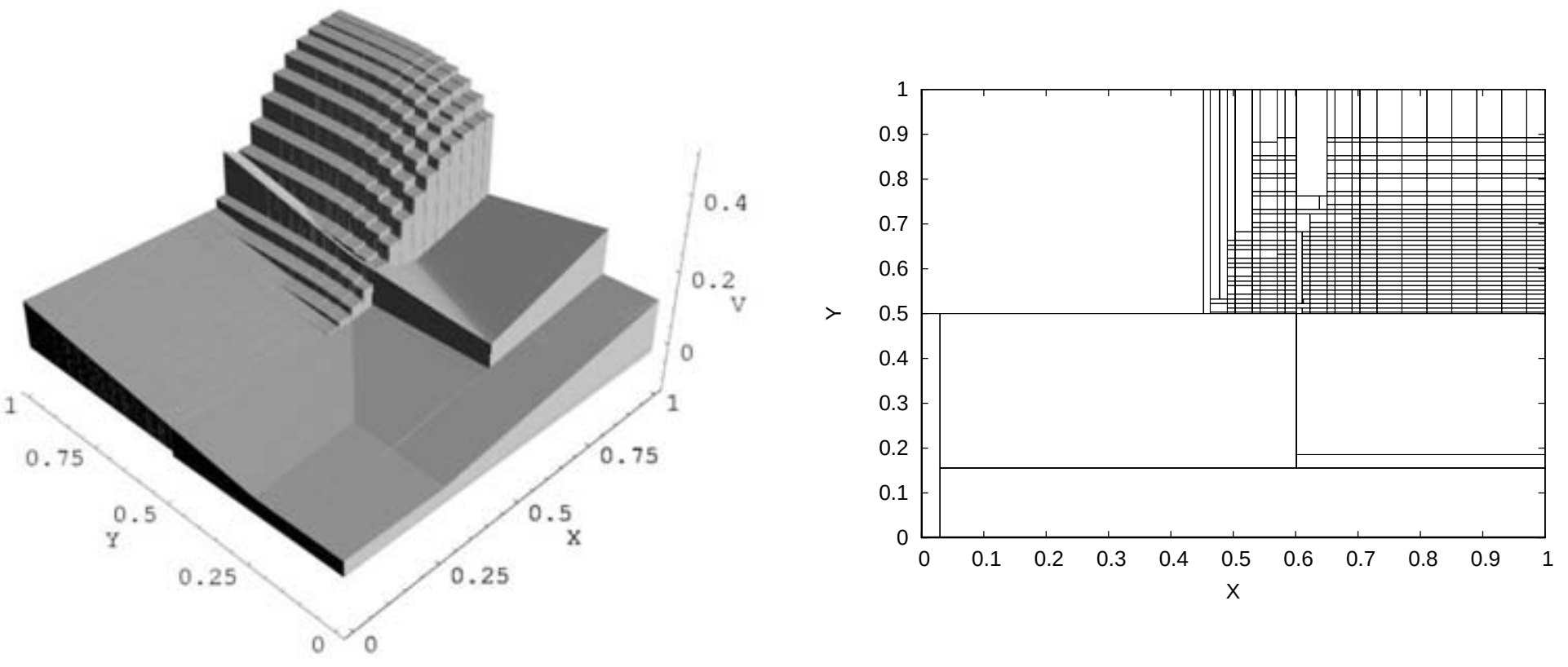

Figure 6: The piecewise-linear value function and the corresponding space partition for the starting discrete state in the 2D problem. The resolution is 25 per continuous variable, to show the discretization more clearly. The left side shows the optimal value function of the starting discrete state computed by the algorithm, and the right side shows the rectangular partition of the continuous space for that value function. For some region, most noticeably the one from (0.6,0.2) to (1.0,0.5), there are multiple linear functions defining the value, resulting in to a curved shape in the left side figure.

Secondly, note the region from around point $(0.6, 0.2)$ to $(1.0, 0.5)$. The value function as can be seen on the left of Figure 6 over this region has a curved shape. It is in fact composed of 13 linear functions. This is typical when the reward model is RPWLC. Again, the dynamic programming is keeping the discretization to a minimal by automatically grouping states whose value function can be represented in a single PWLC form into an abstract state.

For all tests, the solution times for RPWLC models are greater than those of the RPWC model. This is because of the extra computation on the linear vectors in the RPWLC model, in particular, solving the linear programs to keep the representation of the PWLC function minimal. For the 3D problem set, the RPWLC model runs out of memory for the problem with input discretization of 140. The primary cause of this is that some regions require a large number of linear vectors to represent the value function. Our current algorithm attempts to minimize the number of regions. However, we can introduce a trade-off between the size of the partition and the size of the vector representation, by sub-dividing a partition to allow more vectors to be pruned. Handling this trade-off remains as future work.

## 6 RELATED WORK

This paper can be seen as a generalization of both the work on exploiting structure in discrete MDPs (Dietterich & Flann 1997; Boutilier, Dearden, & Goldszmidt 2000; Hoey *et al.* 1999), and the time-dependent MDPs of (Boyan & Littman 2000). The former are restricted to discrete problems but use an analogous structured Bellman backup procedure to ours, while Boyan and Littman's piecewise constant and linear models feature only one continuous state variable. Moving to a multi-dimensional continuous framework introduces many issues in formalism, algorithms, and representation that we have addressed in this paper.

The most common approaches to continuous state variables in MDPs are to use function approximators (Gordon 1995; Boyan & Moore 1995), such as artificial neural networks, to discretize the continuous state space more or less naively, which does not scale well to multiple dimensions, or to use Monte-Carlo approaches (Thrun 2000). None of these approaches exploits the structure in the problem. In fact, we implemented a variant of Thrun's algorithm to compute the function in Figure 1; the algorithm took orders of magnitude longer than the method in this paper. Munos and Moore (Munos & Moore 2002) propose a formal model of a continuous MDP and algorithms for discretizing it adaptively. Their approach involves solving the MDP at one level of discretization, then locally refining the discretization, and repeating until the approximation is good enough. However, their model is a deterministic MDP where actions must be applied over continuous durations. The non-determinism in action outcomes results only from the discretization. This does not fit the problem of planetary rover planning that uses discrete global commands such "drive from lander to rock 1", and is intrinsically rife with exogenous uncertainty (Bresina *et al.* 2002). Although a similar approach based on incremental refinements could probably be adopted in this framework, it is beyond the scope of this paper to explore this direction . Note that by taking advantage of the known structure of the problem, our algorithms find the correct level of discretization and solves the MDP only once.

In the RL literature we find approaches such as U-trees (Uther & Veloso 1998) which learns a tree-based representation of a continuous value function similar to ours. How-ever, since this and similar approaches assume an unknown model, they must infer the value function's structure from observations rather than being able to compute it from the model.

## 7 CONCLUSIONS AND FUTURE WORK

We have proposed an algorithm that exploits structure in the problem representation to solve MDPs with continuous state-spaces (GSSMDPs). We have shown that for restricted subclasses of such MDPs we can solve them exactly, and that for general GSSMDPs we can approximate the model and solve that. By only differentiating states that need to be differentiated, this approach saves considerable computation time, particularly in high dimensions. We then presented the foundations and practical implementation of functional DP algorithms that manipulate piecewise constant and piecewise linear value functions. We then demonstrated our algorithms on simulated Mars rover problems.

The algorithms presented in this paper address only one of the obstacles on the road to applying the decision theoretic approach to real rover problems. There are other issues that considerably limit the size of the problems that can be solved. The main of them is that the lattice of discrete states is itself exponentially large, due to the propositional representation is used. To combat this, we are currently combining the techniques presented here with a structured approach for discrete MDPs in the same line as SPI (Boutilier, Dearden, & Goldszmidt 2000), in an integrated architecture for planning and execution for the K9 experimental rover. At present we maintain a value function over the continuous state for each discrete state. However, it is likely that many of these share structure. In the extreme case, two discrete states may have identical value functions, in which case we would like to combine them. In other cases only a subset of the continuous state may match. We may be able to interleave splits on discrete state with splits on continuous state in the kd-tree to capture the structure in mixed models efficiently.

Another future direction is to allow parameterized actions such as “drive ten metres” or the dawdling action of (Boyan & Littman 2000) that can be performed for any continuous duration. Different semantics for the parameter lead to slightly different models, but in principle there is no reason why our approach cannot be extended to handle actions with continuous parameters.

Finally, we plan to extend our approach to exploit reachability. Similar to (Feng & Hansen 2002), we can perform search using our partitioned state-space and avoid computation completely in regions that are not relevant given a start state.

**Acknowledgements**

We thank Michael Littman for comments on the material. This work was funded by the NASA Intelligent Systems Program. Zhengzhu Feng was supported in part by the National Science Foundation under grants IIS-0219606, and by NASA under cooperative agreement NCC 2-1311. Any opinions, findings, and conclusions or recommendations expressed in this material are those of the authors and do not reflect the views of the NSF or NASA.

## References

Boutilier, C.; Dearden, R.; and Goldszmidt, M. 2000. Stochastic dynamic programming with factored representations. *Artificial Intelligence* 121:49–107.

Boyan, J., and Littman, M. 2000. Exact solutions to time-dependent MDPs. In *Advances in Neural Information Processing (NIPS) 13*. 1–7.

Boyan, J. A., and Moore, A. W. 1995. Generalization in reinforcement learning: Safely approximating the value function. In *Advances in Neural Information Processing (NIPS) 7*, 369–376.

Bresina, J.; Dearden, R.; Meuleau, N.; Ramakrishnan, S.; Smith, D.; and Washington, R. 2002. Planning under continuous time and resource uncertainty: A challenge for AI. In *Proceedings of the 18th Conference on Uncertainty in Artificial Intelligence*.

Cassandra, A.; Littman, M.; and Zhang, N. 1997. Incremental Pruning: A simple, fast, exact method for partially observable Markov decision processes. In *Proceedings of the 13th Conference on Uncertainty in Artificial Intelligence*.

Dean, T., and Givan, R. 1997. Model minimization in markov decision processes. In *Proceedings of the 14th National Conference on Artificial Intelligence*, 106–111.

Dietterich, T., and Flann, N. 1997. Explanation-based learning and reinforcement learning: A unified view. *Machine Learning* 28:169–214.

Feng, Z., and Hansen, E. A. 2002. Symbolic heuristic search for factored markov decision processes. In *Proceedings of the 18th National Conference on Artificial Intelligence*.

Friedman, J.; Bentley, J.; and Finkel, R. 1977. An algorithm for finding best matches in logarithmic expected time. *ACM Trans. Mathematical Software* 3(3):209–226.

Gilks, W.; Richardson, S.; and Spiegelhalter, D. 1996. *Markov Chain Monte Carlo in Practice*. Chapman and Hall.

Gordon, G. J. 1995. Stable function approximation in dynamic programming. In *Proceedings of the 12th International Conference on Machine Learning*, 261–268.

Hoey, J.; St-Aubin, R.; Hu, A.; and Boutilier, C. 1999. SPUDD: Stochastic planning using decision diagrams. In *Proceedings of the 15th Conference on Uncertainty in Artificial Intelligence*.

Munos, R., and Moore, A. 2002. Variable resolution discretization in optimal control. *Machine Learning* 49.

Naylor, B.; Amanatides, J.; and Thibault, W. 1990. Merging BSP trees yields polyhedral set operations. *Computer Graphics (SIGGRAPH'90 Proceedings)* 24(4):115–124.

Thrun, S. 2000. Monte Carlo POMDPs. In *Advances in Neural Information Processing (NIPS) 12*, 1064–1070.

Uther, W. T. B., and Veloso, M. M. 1998. Tree based discretization for continuous state space reinforcement learning. In *Proceedings of the 15th National Conference on Artificial Intelligence*, 769–774.